\documentclass[letterpaper, 10 pt, journal, twoside]{IEEEtran}
\usepackage{lmodern}
\usepackage[T1]{fontenc}
\usepackage{cite}
\usepackage{graphics}    
\usepackage{times}       
\usepackage{amsmath}     
\usepackage{amssymb}     
\usepackage{graphicx}
\usepackage{import}
\usepackage{color}
\usepackage{wrapfig}
\usepackage[font=small,labelfont=bf,skip = 2pt]{caption}
\usepackage[font=small,labelfont=bf,skip = 2pt]{subcaption}
\usepackage{hyperref}
\usepackage{todo}
\usepackage{makecell}
\usepackage{amsmath,bm}

\def\figref#1{Fig.~\ref{#1}}

\begin{document}
\title{Safe Multi-Agent Reinforcement Learning\\for Behavior-Based Cooperative Navigation}

\author{Murad Dawood$^{1,2}$, \and Sicong Pan$^{1}$, \and Nils Dengler$^{1}$, \and Siqi Zhou$^{3}$, \and Angela P. Schoellig$^{3}$ \and Maren Bennewitz$^{1,2}$%
\thanks{Manuscript received: October 1, 2024; Revised February 12, 2025; 
Accepted April 1, 2025.}%
\thanks{This paper was recommended for publication by
Editor Cosimo Della Santina upon evaluation of the Associate Editor and
Reviewers’ comments.
This work has partially been funded and by the Deutsche Forschungsgemeinschaft (DFG, German Research Foundation) under Germany's Excellence Strategy, EXC-2070 -- 390732324 -- Phenorob and under the grant number BE 4420/4-1 (FOR 5351: KI-FOR Automation and Artificial Intelligence for Monitoring and Decision Making in Horticultural Crops, AID4Crops).
 }%

   \thanks{$^{1}$Murad Dawood, Sicong Pan, Nils Dengler and Maren Bennewitz are with the Humanoid Robots Lab, University of Bonn, Germany. 
    {\tt\small dawood@cs.uni-bonn.de, span@uni-bonn.de, dengler@cs.uni-bonn.de, maren@cs.uni-bonn.de}}%
    \thanks{$^{2}$Murad Dawood and Maren Bennewitz are additionally with the Lamarr Institute for Machine
Learning and Artificial Intelligence and the Center for Robotics, Bonn, Germany.}
 \thanks{$^{3}$Siqi Zhou and Angela P. Schoellig are with the Learning Systems and Robotics lab at the Technical University of Munich {\tt\small siqi.zhou@tum.de, angela.schoellig@tum.de}.} 
\thanks{Digital Object Identifier (DOI): see top of this page.}
}%

\markboth{IEEE Robotics and Automation Letters. Preprint Version. Accepted April, 2025}
{Dawood \MakeLowercase{\textit{et al.}}: Safe Behavior-Based Cooperative Navigation} 

\maketitle

\begin{abstract} 
In this paper, we address the problem of behavior-based cooperative navigation of mobile robots using safe multi-agent reinforcement learning~(MARL).
Our work is the first to focus on cooperative navigation without individual reference targets for the robots, using a single target for the formation's centroid. This eliminates the complexities involved in having several path planners to control a team of robots.
To ensure safety, our MARL framework uses model predictive control (MPC) to prevent actions that could lead to collisions during training and execution.
We demonstrate the effectiveness of our method in simulation and on real robots, achieving safe behavior-based cooperative navigation without using individual reference targets, with zero collisions, and faster target reaching compared to baselines.
Finally, we study the impact of MPC safety filters on the learning process, revealing that we achieve faster convergence during training and we show that our approach can be safely deployed on real robots, even during early stages of the training.

\end{abstract}

\begin{IEEEkeywords}
Reinforcement Learning; Robot Safety
\end{IEEEkeywords}
\section{Introduction}
\label{sec:intro}

\IEEEPARstart{C}{ooperative} navigation of unmanned vehicles has gained considerable attention due to its applications in missions such as search and rescue~\cite{liu2016multirobot}, surveillance~\cite{tallamraju2019active}, and payload transfers~\cite{fawcett2023distributed}. As outlined in the literature~\cite{liu2018survey}, cooperative navigation can be achieved through various strategies, including leader-follower, virtual-structures, and behavior-based approaches. Behavior-based methods \cite{xu2014behavior, quan2023robust} offer more flexibility, as the robots' actions adapt to their own observations. The goal in behavior-based cooperative navigation is to maintain relative distances, avoid collisions, and reach target locations.

In this work, we focus on safe learning for behavior-based cooperative navigation. Safety in this context means eliminating all collisions during both training and execution, achieved through safe reinforcement learning (RL). RL has been successful in cooperative navigation~\cite{yan2022relative, sui2020formation, khan2020graph}, but challenges remain due to the sim-to-real gap. This gap arises from real-world sensor noise and unmodeled nonlinearities in simulations. Unlike approaches that focus on improving simulation realism~\cite{rudin2022learning, tobin2017domain}, we emphasize the need to ensure RL is safe for real-world deployment. Although safety in RL has advanced~\cite{zhang2022spatial, sheebaelhamd2021safe, elsayed2021safe}, it remains underexplored in behavior-based navigation. Furthermore, the impact of safety filters on RL training has not been fully studied. In this work, we investigate the effects of applying a model predictive control~(MPC)-based safety filter to RL in the context of behavior-based cooperative navigation.

\begin{figure}[!t] 
\centering \includegraphics[width=0.8\linewidth]{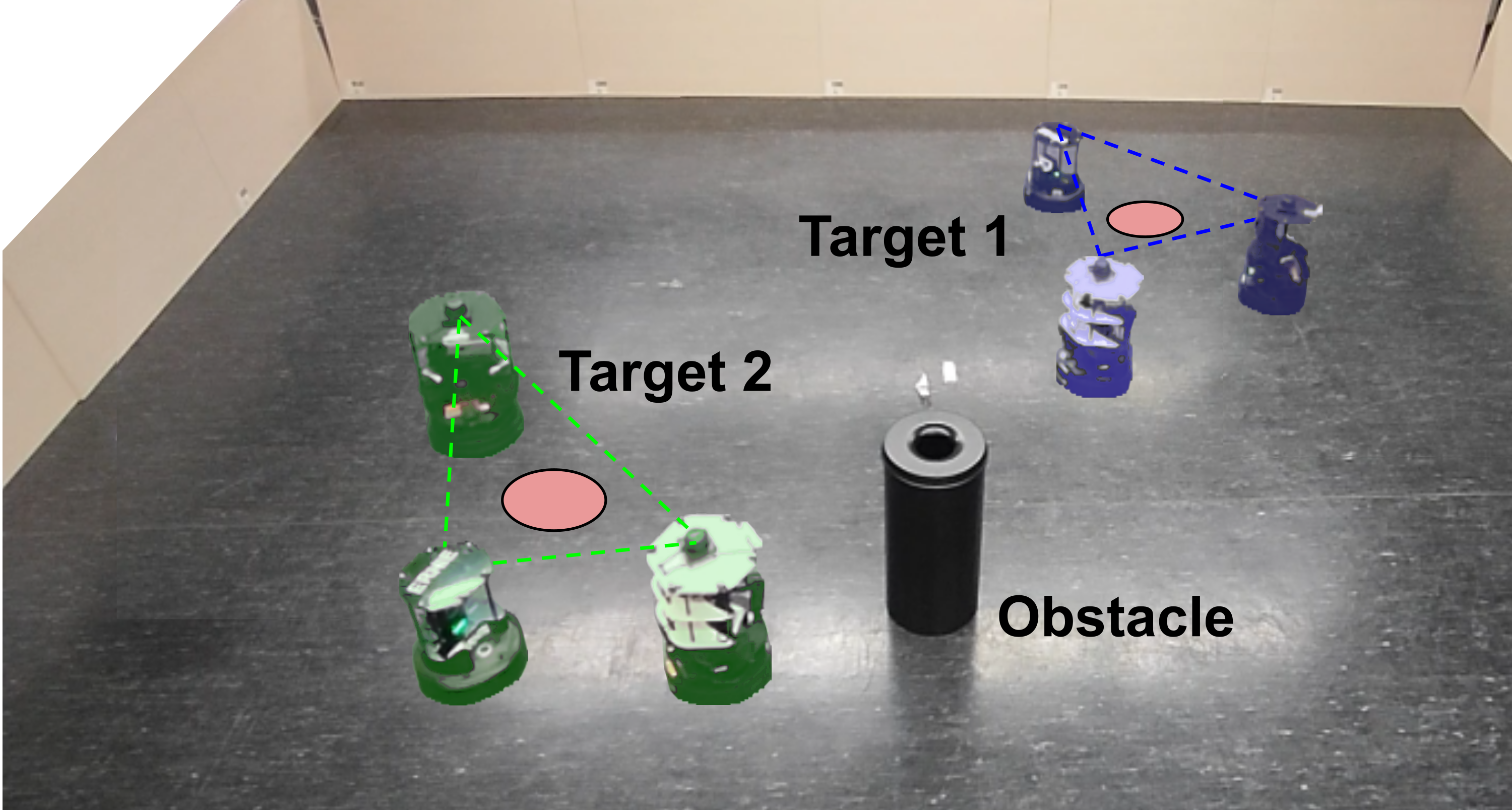}  

\caption{Real-world example for the behavior-based cooperative navigation control.
The robots start from random locations and navigate cooperatively to reach the targets for the centroid of the formation (shown in red) while aiming to maintain the predefined distances with respect to each other.
The blue and green shades show the robots team at the first and second goals, respectively.
} 
\vspace{-20pt}
\label{fig:cover} 
 \end{figure}

To facilitate the deployment of behavior-based cooperative navigation and enable the scalability of our approach, i.e., the application of the learned policy to a team with a higher number of robots without retraining, we use a reference target for the centroid of the formation. This modification is promising for coordinating teams of robots as in the StarCraft multi-agent challenge \cite{samvelyan2019starcraft}.
To maintain the distances between the robots, each robot considers the distances to its two neighbors.
Optimization-based controllers struggle in solving this task, since these controllers require a reference target location for each individual robot and might even assume that each robot has access to all its neighbors positions \cite{xu2014behavior, quan2023robust, zhang2022agile, 10308607, park2023formation, batra2022decentralized, adajania2023amswarmx, quan2022distributed}. 
Therefore, we use multi-agent reinforcement learning~(MARL) to achieve the desired behavior.
Figure~\ref{fig:cover} shows a real-world example of our approach, where the robots get into formation to reach the target locations for the centroid while avoiding unsafe actions.

The primary contributions of this paper can be summarized as follows: \textbf{(i) Safe learning of behavior-based cooperative navigation.
}We introduce the application of safe RL to behavior-based navigation.
This achievement is notable because it ensures the agents' safety during the training phase, a previously unaddressed challenge in behavior-based cooperative navigation, as well as during the execution phase. Additionally, we studied the effect of the safety layer on training efficiency, showing improved convergence compared to baselines.
\textbf{(ii) Eliminating the need for individual path planners for each robot}. We study the behavior of robots in the cooperative navigation task to use less information, resulting in a policy that is adaptable to more robots without the need of retraining. Our approach requires a single reference target for the centroid of the formation to navigate. This setup has not been addressed before in the field of cooperative navigation using MARL so far.
\textbf{(iii) Behavior-based navigation of real robots.
}We demonstrate the performance of the learned policy on real robots.
Employing the distributed MPC safety filters ensures zero collisions throughout the experiments.
Additionally, we show that it is safe to train the policy on the real robots, as our approach eliminates all collisions, even during the early stages of training.
 \vspace{-8pt}
\section{Related Work}
\label{sec:related}
\textbf{Cooperative Navigation Using RL} Several studies applied RL to achieve cooperative navigation using different formulations.
In \cite{yan2022relative} the authors used multi-agents proximal policy optimization\cite{yu2021surprising} along with curriculum learning, reference targets, and relative positions of the robots with respect to each other.
Additionally, \cite{batra2022decentralized} applied PPO for formation control of quadrotors, incorporating individual goals for each unit.
The authors in \cite{sui2020formation} used imitation learning along with RL to formulate a distributed formation controller based on a leader-follower scheme.
In \cite{khan2020graph} the authors used policy gradients along with a graph convolutional network~(GCN) and reference targets for each robot to achieve static formations of drones.
\cite{xie2021reinforcement} and \cite{zhou2019learn} achieved formation control of maritime unmanned surface vehicles using deep deterministic policy gradient (DDPG) and deep Q-networks, respectively, based on leader-follower approaches. \textcolor{black}{Moreover, \cite{he2022multiagent} proposed a motion planner based on a multi-agent extension of the soft-actor-critic~(SAC) to coordinate multiple robots to reach their respective goals.}
\textbf{Different from the previous works}, we only use reference targets for the centroid of the formation and information about only two neighbors for each robot.
More importantly, we ensure collision-free training by using MPC-based safety filters, and we use the attention mechanism to account for the interaction between the robots.

\textbf{Safety in MARL:} Safety is a crucial aspect in MARL, with ongoing developments addressing this concern. \cite{cai2021safe} utilized control barrier functions (CBF) \cite{ames2019control} along with multi-agent DDPG (MADDPG) for collision-free navigation of two agents. \cite{zhang2022spatial} explored the use of attention modules and decentralized safety shields based on CBF to reduce collisions in autonomous vehicle scenarios. Similarly, \cite{sheebaelhamd2021safe} implemented a centralized neural network as a safety layer with MADDPG in cooperative navigation. Moreover, \cite{zhang2019mamps}  proposed using a predictive shielding approach, along with MADDPG,
to ensure the safety of multi-agents in cooperative navigation scenarios. \cite{zhang2022barrier} presented RL-based model predictive control to achieve
leader-follower formation control of mobile robots while ensuring their safety. \textbf{In contrast to the
previous works}, we focus on behavior-based navigation where collisions between agents during
navigation \textcolor{black}{ are more likely to occur}, since the agents are required to navigate close to each other
and not just have a single instance where their paths intersect.
Additionally, the previous studies focused on the task of having individual reference targets, which can be achieved by training a single RL policy and then deploy it to several robots as in \cite{brito2021go, vinod2022safe,li2022decentralized}.
Our task, on the other hand, necessitates using a MARL framework so that the robots interact with each other during the training to rely only on a reference location for the centroid.
While the previous studies fell short of real-world experiments, we conducted extensive experiments with both trained and untrained policies on real robots to demonstrate the safety of our framework, and the performance of  our behavior-based modification.


\begin{figure}[!t] 
\centering
	 \includegraphics[width=0.9\linewidth]{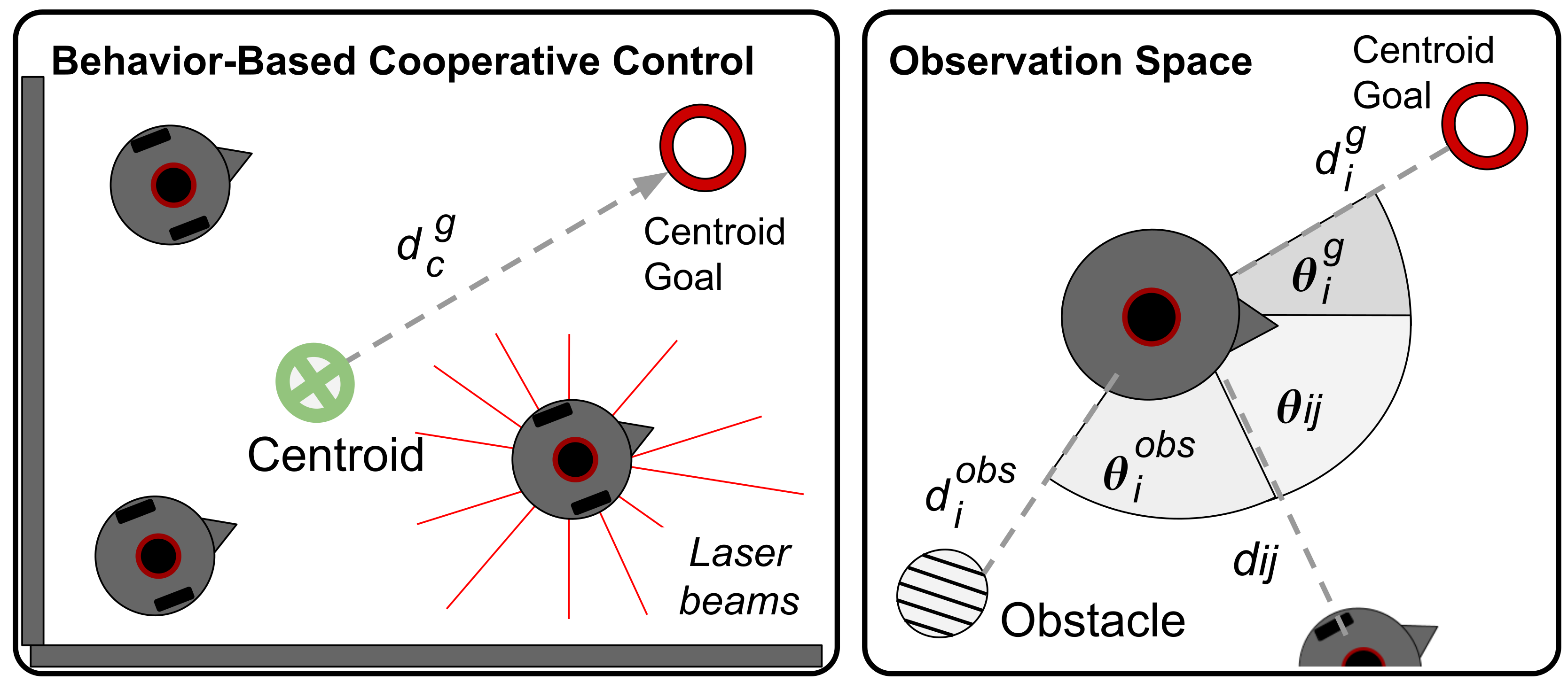}

\caption{The observation space per robot includes lidar readings (red lines), distances and headings to the goal ($d_{i}^{g}$, $\theta_{i}^{g}$), two neighbors ($d_{ij}$, $\theta_{ij}$), and the closest obstacle ($d_{i}^{obs}$, $\theta_{i}^{obs}$). Additionally, robots have information about the centroid's distance to the goal ($d_{c}^{g}$) $d_{c}^{g}$.} 
\label{fig:formation}
\vspace{-15pt}
 \end{figure} 
 
\vspace{-5pt}
\section{Problem Statement}
\label{sec:prob}
\vspace{-5pt}
We consider the task of safely training a team of mobile robots to move the centroid of their formation to a desired target location while avoiding collisions with static obstacles and neighboring robots, and aiming to maintain predefined distances between robots.
We assume that each robot is equipped with a lidar sensor to perceive its surroundings, each robot can localize itself within the environment, and that there is no pre-knowledge about the environment or existing obstacles. To ensure the safety of the robots at all times, each robot has its own safety layer to avoid collisions with its neighbors and the obstacles. The robots do not communicate with each other, but navigate cooperatively based on information about their centroid and the reference target, and the relative distances. \textcolor{black} { Each robot receives relative distances and angles only about its two neighbors even if the team consists of more robots, as well as the target location of the centroid, as illustrated in Fig. \ref{fig:formation}. Finally, each robot calculates the distance and relative angle to the closest obstacle from the lidar scan.} We do not assume that the global information is available for each robot. This enables the scalability of our approach when applied to more robots, without retraining the policy.

\begin{figure}[!t] \centering \begin{subfigure}{0.75\linewidth} 
\centering \includegraphics[width=\linewidth]{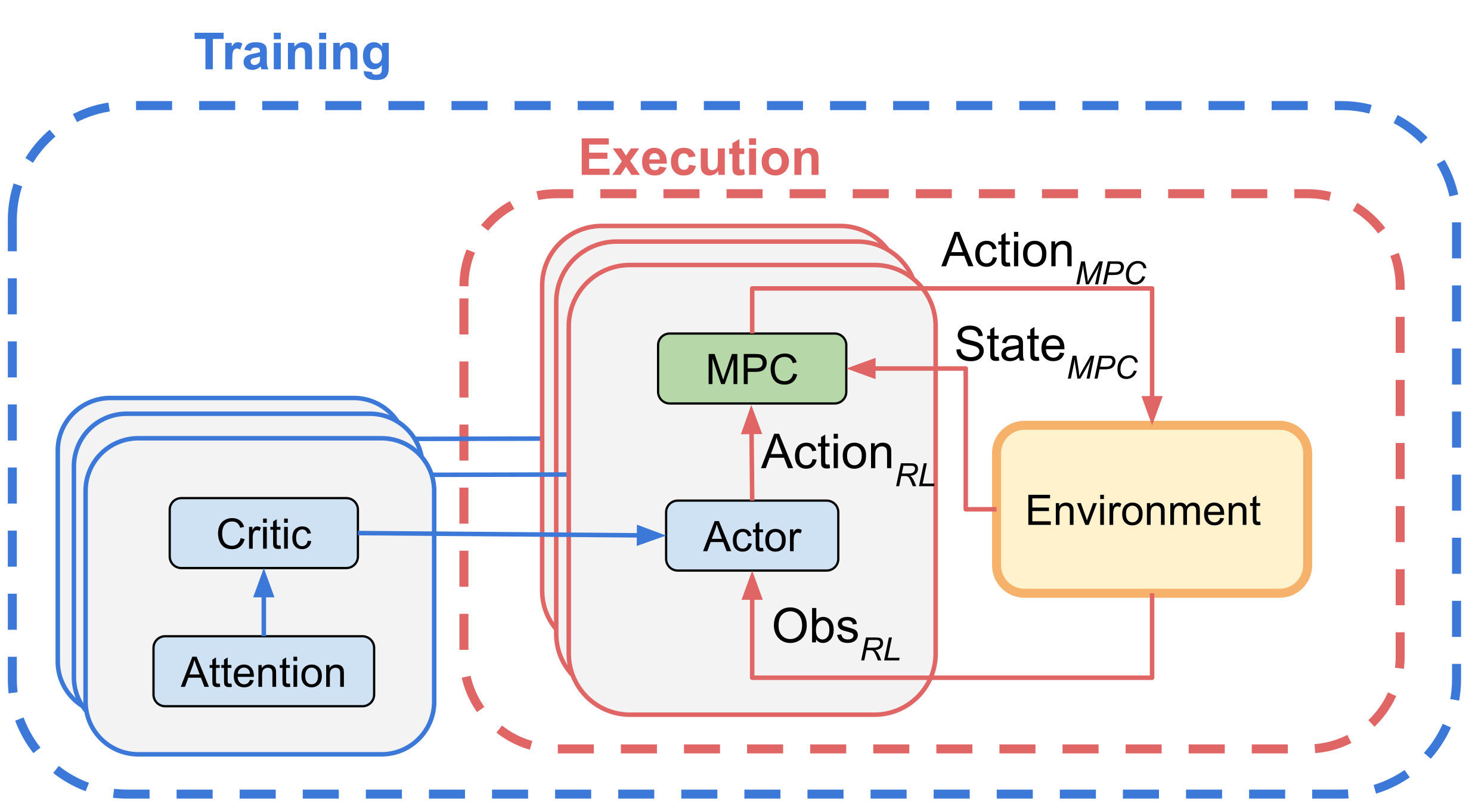} \subcaption{Centralized training decentralized execution (CTDE)} \label{fig:architecture} \end{subfigure} \begin{subfigure}{0.75\linewidth} \centering \includegraphics[width=\linewidth]{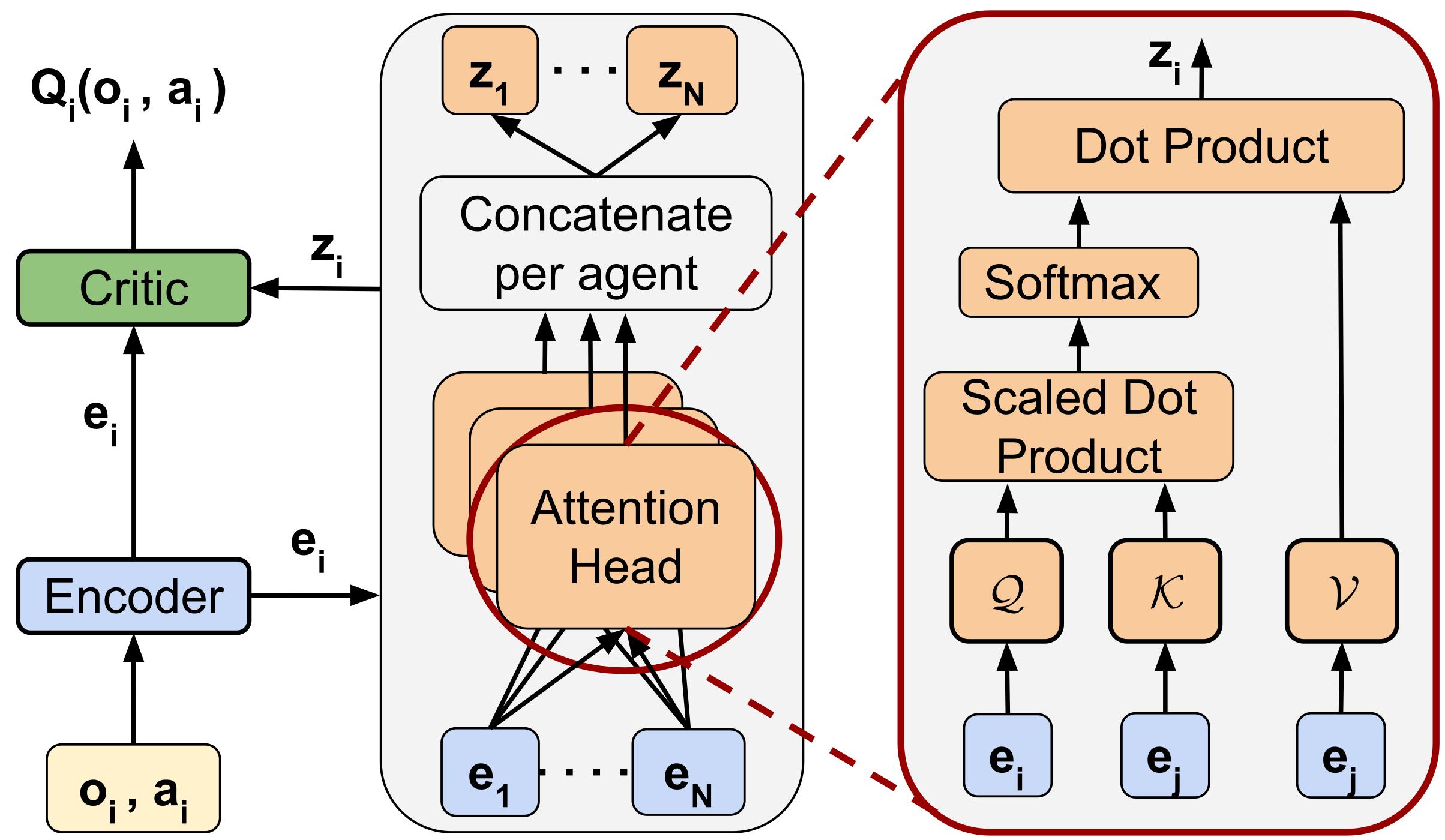} \subcaption{Attention-based multi-agent reinforcement learning}\label{fig:att1} \end{subfigure} 

\caption{Illustrations of the CTDE architecture and attention module.
\textbf{(a)} At each time step, the agent interacts with the environment to receive the current observation $Obs_{RL}$, including relative information about the two neighbors and the closest obstacle, and outputs the Action$_{RL}$. The MPC controller receives the State$_{\mathit{MPC}}$, overrides any unsafe actions to prevent collisions, and sends Action$_{\mathit{MPC}}$ to the robot. During training (blue dashed), critics share all agents' observations, while during execution (red dashed), each actor accesses only its own observation.
\textbf{(b)} In the attention-based critics, observations are encoded separately and fed into attention heads to calculate weights based on the query, and the key-value pairs. The attention output is then concatenated with the states and fed into the critics.
}  
\vspace{-21pt}
\end{figure} 

\vspace{-5pt}
\section{Our Approach}
\label{sec:app}
\vspace{-5pt}
In this section, we formulate our work as a MARL problem, introduce the attention module for agent interaction, and describe the MPC-based distributed safety filters in detail.

\vspace{-4pt}
\subsection{Multi-Agent Reinforcement Learning (MARL): }In this work, we adopt the Markov games' framework~\cite{littman1994markov}, which is formulated as a set of partial observable Markov decision processes~(POMDPs).
For the case of $N$ agents, we have a set $S$ that represents the augmented state space of all the agents, a set of actions $A_{1},...,A_{N}$ and a set of observations $O_{1},...,O_{N}$ for each agent. At each time step $t$, an agent $i$, receives an observation correlated with the state $o^{t}_{i}: S \rightarrow O_{i}$, which contains partial information from the global state $s^{t} \in S$. Where $s^{t}$ refers to $state_{RL}$ in Fig. \ref{fig:architecture} . The agent chooses an action $a^{t}_{i}$ according to a policy, $\pi_{i} : O_{i}\rightarrow  A_{i}$, which maps each agent's observation to an action. The agent then receives a reward as a function of the state and the action taken, $r_i : S \times A_{i} \rightarrow \mathbb{R}$. For each agent, the tuple containing the state, the action, the reward, and the next state is added to the respective replay buffer $\mathcal{D}_{i}$. 

Our approach is based on the soft-actor-critic~(SAC)~\cite{haarnoja2018soft} algorithm.
We extend SAC to the multi-agent domain and apply the centralized-training decentralized execution~(CTDE)~\cite{foerster2016learning} scheme.
During \textit{training} all the agents share their observations with each other.
However, during \textit{execution}, each agent has access to its own observation only as shown in Fig.
\ref{fig:architecture}.

For each agent we define the \textbf{observation space} as in Fig.~\ref{fig:formation}.
\textcolor{black}{The observation includes 40 equiangular lidar readings, the relative distances $d_{i}^{j}$ and angles $\theta_{i}^{j}$ to its two neighbors, the relative distance $d_{i}^{goal}$ and angle $\theta_{i}^{goal}$ to the goal of the centroid, the distance from the centroid of the formation to the goal $d_{centroid}^{goal}$, the relative distance $d_{i}^{obs}$ and angle $\theta_{i}^{obs}$ to the nearest obstacle (calculated from the lidar scan), the desired reference distance between the robots,} and the actions of the robot for the previous time step $a^{t-1}_{i}$ which is the pair ($v^{t-1}_{i}, w^{t-1}_{i}$).
For the \textbf{action space}, we control the linear and angular velocities $v$ and $w$.

The \textbf{reward} function for each robot $i$ is defined as:
 \vspace{-2pt}
\[
\begin{aligned}
\mathbf{r^{t}_{i}} = & \ \mathbf{r_{\mathit{goal}} \cdot \mathbf{1}_{\text{goal reached}} + r_{\mathit{i}}^{\mathit{collision}} \cdot \mathbf{1}_{\text{collision or stuck}}} \\
& + \mathbf{(r_{\mathit{i}}^{\mathit{formation}} + r_{\mathit{i}}^{\mathit{obs}} + r_{\mathit{centroid}}^{\mathit{goal}}) \cdot \mathbf{1}_{\text{otherwise}}}
\end{aligned}
\]
 \vspace{-2pt}
where \(\bm{r_{\mathit{goal}}}\) and \(\bm{r_{\mathit{i}}^{\mathit{collision}}}\) are sparse rewards indicating whether the centroid reaches the target and whether the robot collides or remains stuck for several steps, respectively. 
\(\bm{r_{\mathit{i}}^{\mathit{formation}}}\) is a reward for maintaining formation, computed as the negative error between the reference distances and actual distances among agents. 
\(\bm{r_{\mathit{i}}^{\mathit{obs}}}\) is a reward for maintaining a safe distance from the closest obstacle, given by the negative obstacle distance. 
\(\bm{r_{\mathit{centroid}}^{\mathit{goal}}}\) is a reward for guiding agents toward the goal, calculated as the negative distance between the centroid and the goal.

\vspace{-8pt}
\subsection{Attention-Based Critics:} To effectively capture relative information among agents, we implement an attention module (\figref{fig:att1}) following the framework of \cite{vaswani2017attention}, which computes attention based on queries, keys, and values. Each agent's observation is encoded into embeddings \(e_{1} \ldots e_{N}\) by an encoder network. These embeddings are input to the \(K\) and \(V\) networks to generate keys and values, respectively, while the querying agent's embedding is processed by the \(Q\) network. The computed attention weights are then used to enhance the encoded observations, which are concatenated with the outputs \( \mathsf{z_{i}} \) from the attention module and input into the critics to compute Q-values.

\begin{figure*}[!t] 
\centering 
\begin{subfigure} {0.245\linewidth} \includegraphics[width=\linewidth]{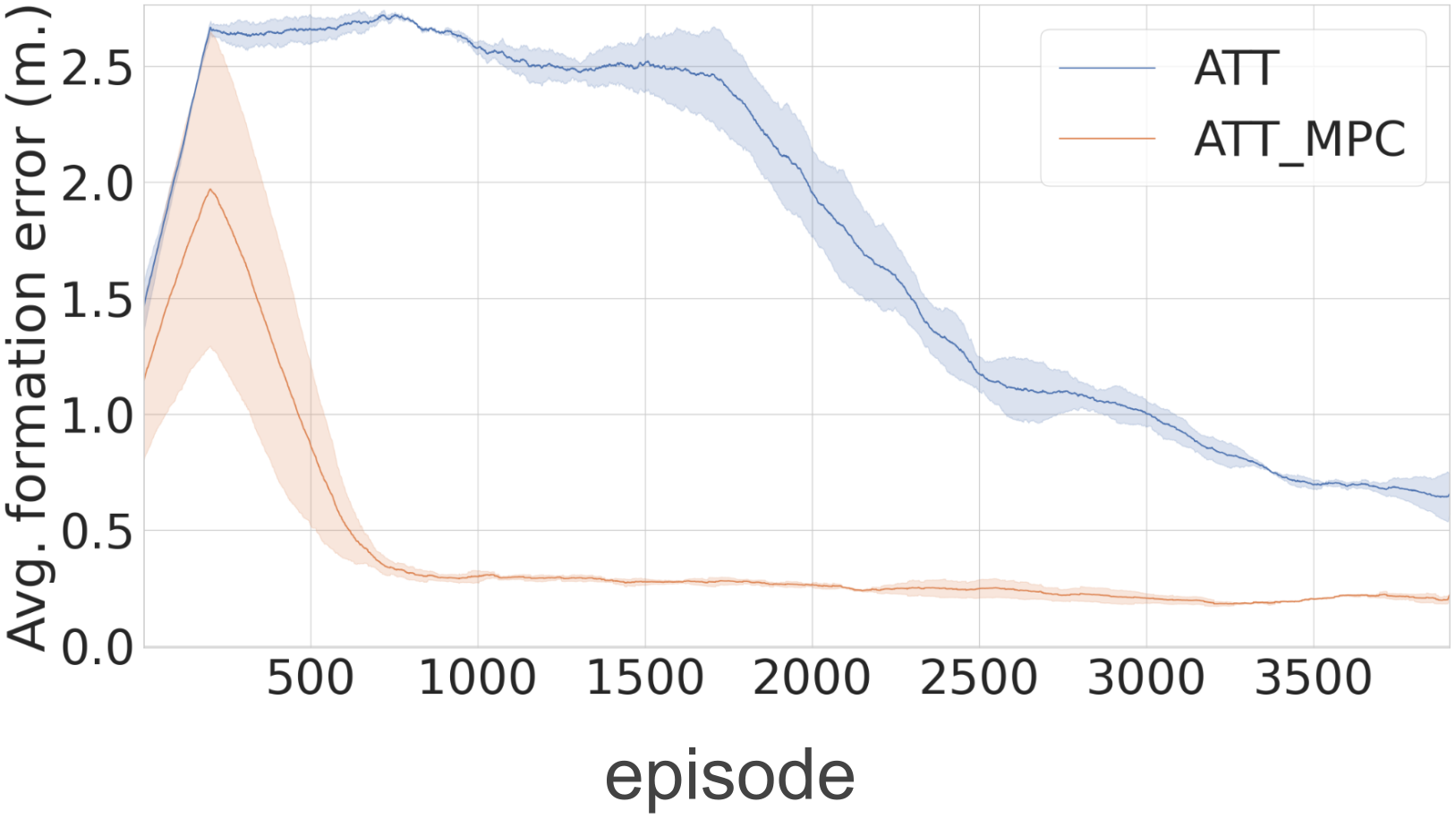} \subcaption{\textbf{Average formation error} during the \textbf{first} environment.} \label{fig:form_mpc} \end{subfigure} 
\begin{subfigure} {0.245\linewidth} \centering 	\includegraphics[width=\linewidth]{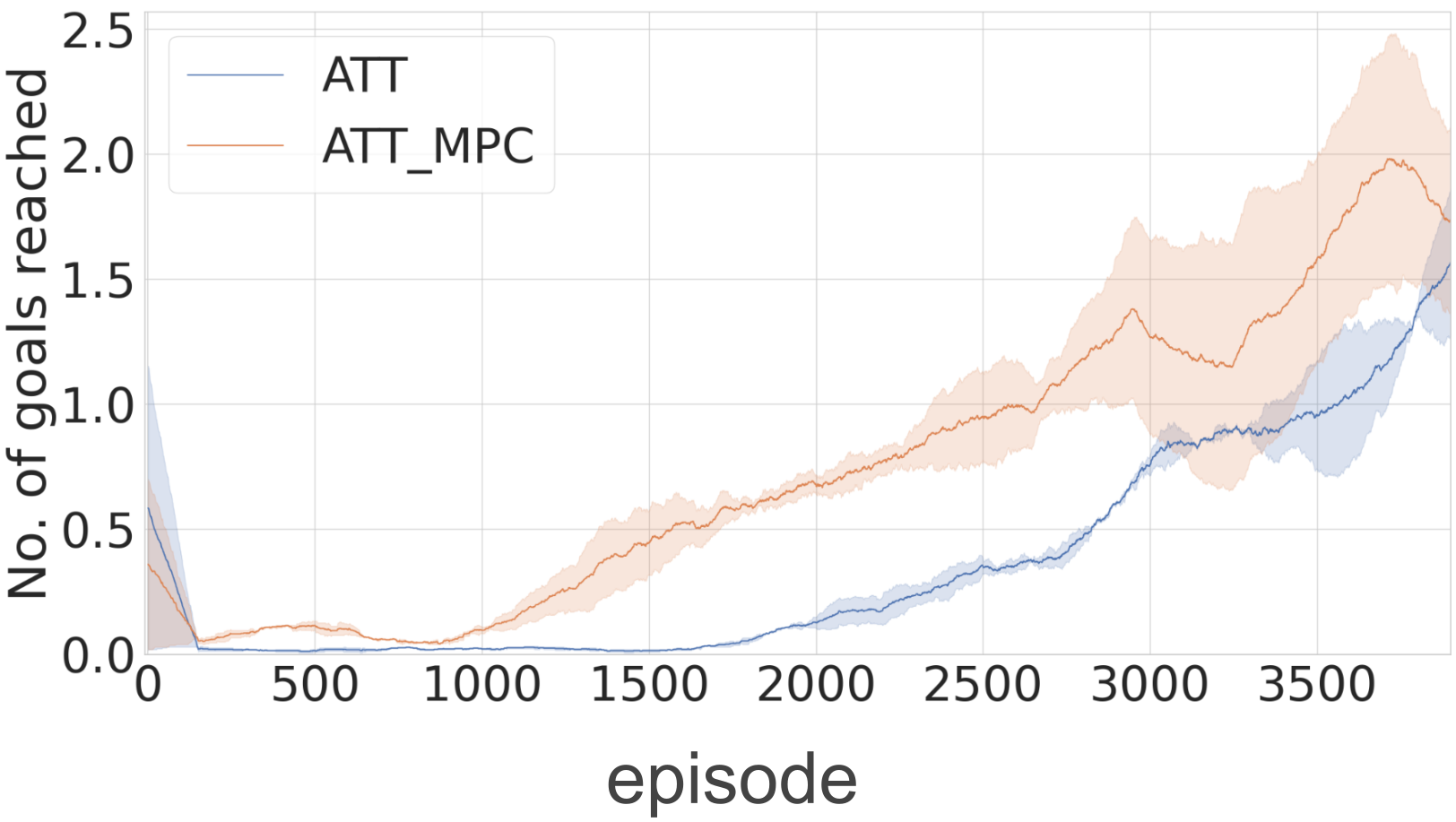} \subcaption{\textbf{Number of goals} reached during the \textbf{first} environment.} \label{fig:goals}  \end{subfigure} 
\begin{subfigure} {0.245\linewidth} 
\includegraphics[width=\linewidth]{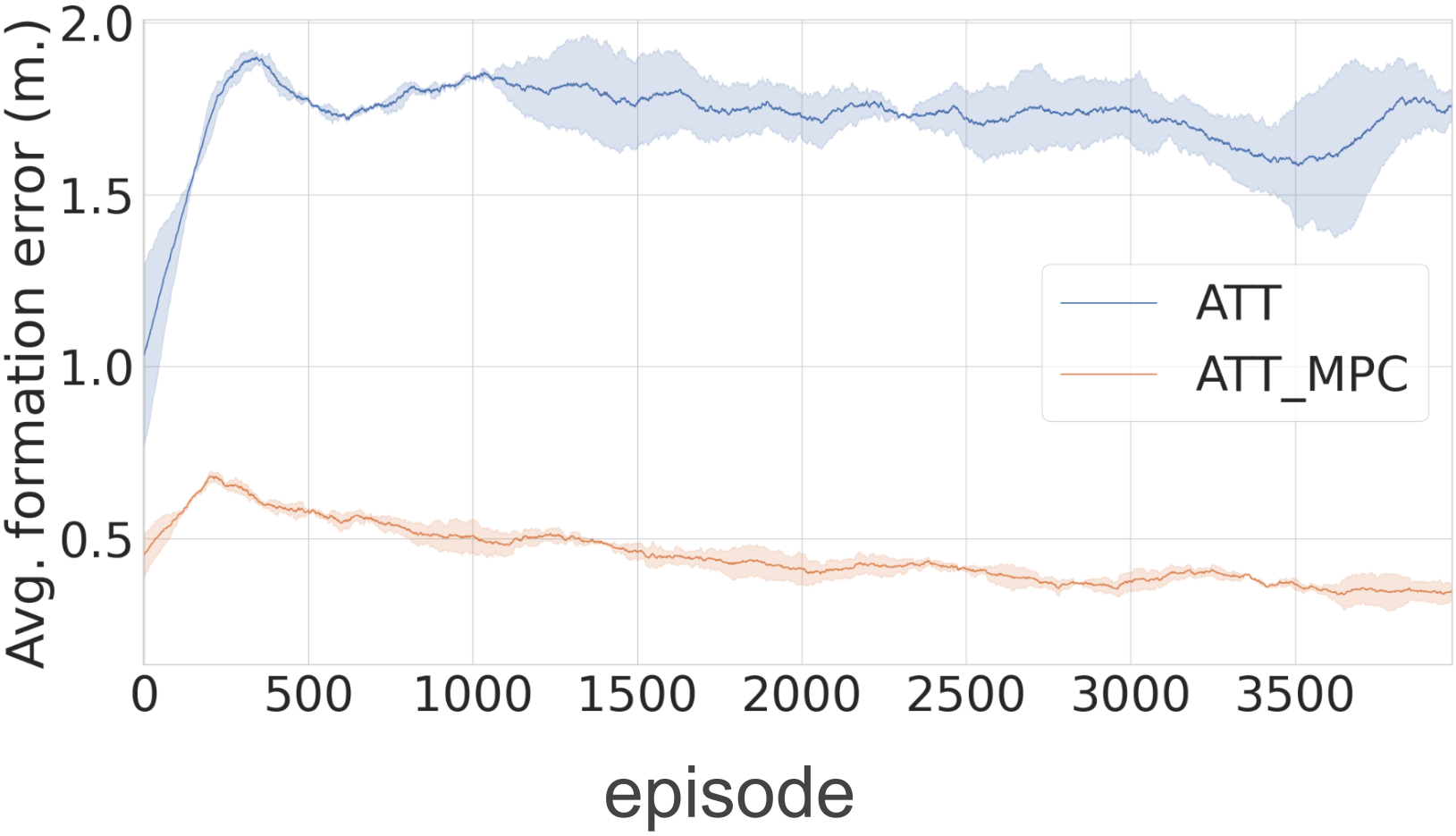} 
\subcaption{\textbf{Average formation error} during the \textbf{second} environment.} \label{fig:form2} \end{subfigure} \begin{subfigure} {0.245\linewidth} 
\centering 	
\includegraphics[width=\linewidth]{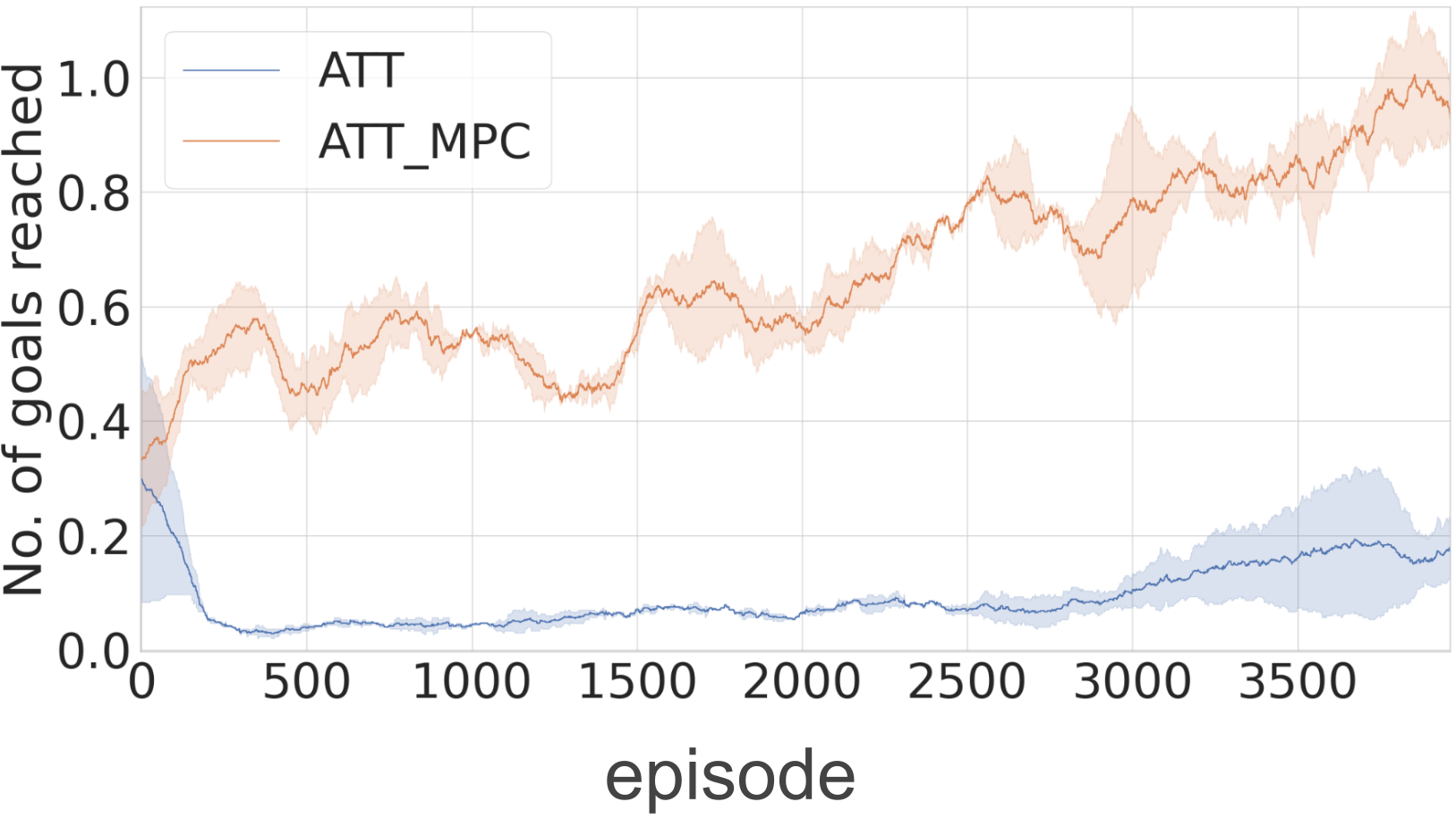} 
\subcaption{\textbf{Number of goals} reached during the \textbf{second} environment.} \label{fig:goals2} \end{subfigure} 
     
\caption{
The figures show the impact of the safety filter on training, with bold lines representing the average and shaded areas indicating the standard deviation across three random seeds. \textbf{(a,c)} depict the average formation error in the first and second training environments, while \textbf{(b,d)} display the number of goals achieved per episode, in both environments. The agent with the MPC (\textbf{ATT\_{MPC}}) consistently surpasses the pure learning agent (\textbf{ATT}) in reducing formation errors and increasing goal achievements. This demonstrates that incorporating a safety filter reduces the number of episodes required to achieve the desired performance compared to pure learning agents.
} 
 \label{fig:filter}
 \vspace{-17pt}
 \end{figure*} 
\vspace{-8pt}
\subsection{Model Predictive Safety Filter:} To ensure the safety of the agents, we implement distributed shields based on nonlinear model predictive control (NMPC).
The MPC utilizes a mathematical model of the robot (\ref{eq:model}) to calculate the predicted states based on the current state and action.
At each time step, the MPC solves an optimization problem (\ref{eq:mpc}) to find the optimal control sequence that satisfies a set of predefined constraints on the states and actions.
Only the first action of the control sequence is applied while the rest of the sequence is discarded.
Prior safety related studies have employed the MPC as a safety filter \cite{wabersich2021probabilistic, zhang2019mamps, bejarano2023multi} due to its capability to handle systems with multiple states, controls, and constraints. To minimize the intervention of the MPC-safety filter and avoid disrupting the exploration of the RL agent, the mentioned studies align the MPC actions with those of the RL agent. In our work, we additionally, penalize the RL agent for deviating from the MPC safe actions, which in turn teaches the RL agent to not rely on the MPC-safety filter as we show in Sec.\ref{sec:ours_no_mpc}.

\subsubsection{Safety Filter Formulation} The robot's state in NMPC ($State_\mathit{MPC}$ in Fig. \ref{fig:architecture}) is defined as $\textbf{x} = [x, y, \theta]^{T}$, representing its position and heading. Controls are denoted by $\textbf{a} = [v, \omega]^{T}$, matching the RL actions. A weighted Euclidean norm with
a positive definite weighting matrix $R$ is denoted as $||\textbf{x}||^2_{R} = \textbf{x}^{T}R\textbf{x}$.

\subsubsection{Prediction Model} The discrete-time model of the robot is:
\begin{equation}
 \textcolor{black}{
\small
\textbf{x}_{t+1} =  f(\textbf{x}_{t}, \textbf{a}_{t}) = \textbf{x}_{t} + \begin{bmatrix}
cos \theta_{t} & 0\\
sin \theta_{t} & 0\\
0 & 1
\end{bmatrix} \textbf{a}_{t} \Delta t}
\label{eq:model}
\end{equation}
\subsubsection{Optimal Control Problem} To ensure the safety of the robot while minimizing the intervention of the safety filter, we formulate the optimization problem for each robot $i$ at time step $t$ as follows:

\begin{subequations} {
\vspace{-15pt}
\small
\begin{align}
 \underset{\substack{\textbf{x}_{t:t+T|t},\\ \textbf{a}_{t:t+T-1|t}}}
 {\mathrm{min}}
& \lVert \textbf{a}_{\mathit{RL}} - \textbf{a}_{t|t} \rVert^2_{\mathit{R_{0}}} + \sum_{k=1}^{T-1}  \lVert \textbf{a}_{t+k|t} \rVert^2_{\mathit{R}} + \sum_{k=0}^T  \lVert \dfrac{1}{e^{{\mathit{dist}}^{t+k|t}_{\mathit{i,1}}}} \rVert^2_\mathit{D} \nonumber\\
 &+ \sum_{k=0}^T \lVert\dfrac{1}{e^{{\mathit{dist}}^{t+k|t}_{\mathit{i,2}}}} \rVert^2_\mathit{D} + \sum_{k=0}^T \lVert \dfrac{1}{e^{{\mathit{dist}}^{t+k|t}_{\mathit{obst}}}} \rVert^2_\mathit{D}  \label{eq:cost}\\
 \text{s.t.} \quad 
& \textbf{x}_{t|t} = \textbf{x}_{t},\label{eq:co1}\\
 & \textbf{x}_{t+k+1|t} = f(\textbf{x}_{t+k|t}, \textbf{a}_{t+k|t}), \forall k= 0,1,...,T-1 \label{eq:co2}\\
 &  \textbf{a}_{t+k|t} \in \textbf{A},\hspace{2.23cm} \forall k= 0,1,...,T-1 \label{eq:co3}\\
&  \textbf{x}_{t+k|t} \in \textbf{X},\hspace{2.23cm} \forall k= 0,1,...,T-1, \label{eq:co4}
\end{align}}\label{eq:mpc}
\vspace{-18pt}

\end{subequations}
where $T$ represents the prediction horizon. The notation ${t+k|t}$ indicates predictions at time $t+k$, assuming the current time is $t$. The optimization terms in (\ref{eq:cost}) are structured in three main components.
 The first term measures the deviation between the RL agent's proposed action $\textbf{a}_{\mathit{RL}}$ (Action$_{RL}$ in Fig. \ref{fig:architecture}) and the initial MPC action $\textbf{a}_{0}$ (Action$_{\mathit{MPC}}$), optimizing for minimal discrepancy.
The second term minimizes the magnitude of future control signals ${\textbf{a}}_{\mathit{k}}$, promoting smoother transitions.
Third, the distances ${\mathit{dist}}_{\mathit{i,1}}$ and ${\mathit{dist}}_{\mathit{i,2}}$ between the robot and its two neighbors, as well as ${\mathit{dist}}_{\mathit{obst}}$ to the nearest obstacle, are penalized to ensure safety.
The exponential function applied to distances has proven effective in enhancing obstacle avoidance, as shown in previous research~\cite{dawood2023handling}. \textcolor{black}{ The weight matrices $R_{0}$, $R$, and $D$ are tuned such that $R_{0}$ has higher weights than $R$ to ensure that the first action matches the proposed action $\textbf{a}_{\mathit{RL}}$. $D$ has the highest weight to maximize the cost on getting close to obstacles.}

Constraints (\hyperref[eq:co1]{\ref*{eq:co1}--\ref*{eq:co4}}) ensure adherence to system dynamics and operational limits, defined as follows:
 (\ref{eq:co1}) sets the initial state,
  \textcolor{black}{(\ref{eq:co2}) enforces the robot's dynamic model from Eq. (\ref{eq:model})},
 (\ref{eq:co3}) and (\ref{eq:co4}) maintain the actions and states within predefined bounds, where $\textbf{X}$ and $\textbf{A}$ denote the allowable sets of states and controls, respectively.

The safety filter, operating independently of the behavior-based navigation task, ensures robot safety by aligning with RL agent actions to maintain exploratory integrity during training. It is compatible with any MARL framework as it does not change the RL algorithm as we show in Sec.\ref{sec:baselines_mpc}. 
For implementation, we employed acados~\cite{Verschueren2019}, a tool for solving nonlinear optimal control problems.

\vspace{-7pt}
\section{Experimental Evaluation}
\label{sec:exp}
The main focus of this work is to achieve behavior-based navigation of mobile robots while ensuring the safety of the robots.
We design the experiments so that we can: \textbf{(i)} study the effect of using the MPC on the training of the RL agents, showing that we can achieve collision free training to learn the task in fewer number of episodes (Sec.~\ref{sec:mpc_filter}), \textbf{(ii)} compare our approach against baselines in simulations, demonstrating that our approach outperforms state-of-the-art baselines, that our RL agent learns to not rely on the MPC-safety layer indefinitely, and that the MPC-safety layer can be integrated with other RL algorithms (Sec.~\ref{sec:sims}:\ref{sec:baselines_mpc}), \textbf{(iii)} show that our approach can be transferred on real robots (Sec.
\ref{sec:real}), \textbf{(iv)} test the ability of the trained policy to generalize to more agents than used during training (Sec. \ref{sec:gener}).

\textcolor{black}{\textbf{Training Details:} For the attention module, we used four attention heads and a hidden dimension of 128. For our approach and the RL baselines, we employed MLPs with two hidden layers of size (256, 256) for both the critics and the actors, using ReLU activation functions. The entropy coefficient in SAC is optimized online. The learning rate for all modules is set to $3 \times 10^{-4}$, with a batch size of 128 and a replay buffer size of $1 \times 10^{6}$. The observation space and reward are fixed for all RL approaches. In simulations, we used the Turtlebot3 (TB3) robots, while in the real world experiments we used two TB2 robots and one TB4 robot.}

\begin{table*}[!t]
\centering
\resizebox{17cm}{!}
{
\begin{tabular}{|c|c|c|c|c|c|c|c|c|c|c|}
\hline
 & \multicolumn{3}{c|}{\textbf{Goals Reached W/o Obs.}} & \multicolumn{3}{c|}{\textbf{Goals Reached W/ Obs.}} &\multicolumn{4}{c|}{\textbf{S-Shaped Path}} \\
\hline
\textbf{Agent} & \textbf{Success \textuparrow}  & \textbf{Colls \textdownarrow} & \textbf{Tout \textdownarrow} & \textbf{Success \textuparrow} & \textbf{Colls \textdownarrow} & \textbf{Tout \textdownarrow} &  \makecell{\textbf{Goals Reached \textuparrow}} & \makecell{\textbf{Time (sec.) \textdownarrow}} &\makecell{\textbf{Colls \textdownarrow}} &\hfil \makecell{\textbf{Avg form err (m) \textdownarrow}} \\ 
\hline
Ours & \textbf{99.5\%} & \textbf{0} & \textbf{0.5\%} & \textbf{96.2\%} & \textbf{0} & \textbf{3.8\%} &  \textbf{100\%} & \hfil \textbf{108.93 $\pm$ 14.8} &  \textbf{0} & \hfil 0.392 $\pm$ 0.17\\
\hline
MASAC \cite{he2022multiagent} & 58\% & 8.9\% & 33.1\% & 51.8\% & 26.3\% & 21.9\% & 75.8\% & \hfil 134.63 $\pm$ 21.3 & \hfil 2.8 $\pm$ 3.65 & \hfil 1.25 $\pm$ 0.15\\
\hline
MADDPG \cite{lowe2017multi} & 9.5\% & 57.7\% & 32.8\% & 6.6\% & 81.5\% & 11.9\% & 12.4\% & \hfil 118.9 $\pm$ 28.39 & \hfil 4.16 $\pm$ 3.81 & \hfil 4.05 $\pm$ 4.06\\
\hline
DMPC \cite{dawood2023handling} & 13.9\% & 0 & 86.1\% & 6.9\% & 0 & 93.1\% & 13.4\% & \hfil  162.7 $\pm$ 15.53 & 0 & \hfil\textbf{ 0.127 $\pm$  0.184}\\
\hline

& \multicolumn{10}{c|}{\textbf{ Can we execute our method without the MPC?}}\\
\hline
\makecell{Ours without MPC} & 98.6\% & 1.4\% & 0 &  96\%  & 2.5\% & \textbf{1.5\%}  & 93.7\% &  112.35 $\pm$  16.7&  0.45 $\pm$ 1.48 & \hfil  0.42 $\pm$  0.23\\

\hline
& \multicolumn{10}{c|}{\textbf{Integrating the MPC Safety Layer with the Baselines}}\\
\hline
\makecell{MASAC \cite{he2022multiagent} (With MPC)} & 67.7\% & 0\ & 32.3\% &  60.3\%  & 0\ & 39.7\% & \hfil 79.9\% & \hfil 129.54 $\pm$ 38.7 & \hfil 0  & \hfil 0.98 $\pm$ 0.18\\
\hline
\makecell{ MADDPG \cite{lowe2017multi} (With MPC)} & 10.4\% & 0\ & 89.6 \% &  7.6\%  & 0\ &  92.4\% & \hfil 13.05\% & \hfil 164.9 $\pm$ 41.02 & \hfil 0 & \hfil 3.56 $\pm$ 3.67\\
\hline
\end{tabular} }
\caption{The table shows the performance of our approach against multi-agent SAC~(MASAC), multi-agent DDPG~(MADDPG), and decentralized model predictive control~(DMPC) in simulation. Additionally, it demonstrates the results for testing our approach without the MPC-safety filter, and testing the baselines with the MPC-safety filter. The percentage reflects the proportion of achieved goals, collisions, and timeouts in relation to the 1,000 trials in case of the goal reaching scenarios, and 120 tests for the S-shaped path test. Our approach makes zero collisions in all configurations, and takes less time to reach the desired locations. Disabling the MPC-filter results in fewer successful runs in all scenarios, as collisions tend to happen. Eliminating the collisions from the MASAC results in better performance with respect to reaching more goals. For the MADDPG, there is no noticeable improvement.}
 \vspace{-13pt}
\label{tab:sims}
\end{table*}

\vspace{-10pt}
\subsection{Training With the MPC Filter}
\label{sec:mpc_filter}
\begin{table}[!t]
\centering
\resizebox{6.5cm}{!}{
\begin{tabular}{|c|c|c|}
\hline
\textbf{No. of Obstacles} & \textbf{Success\textuparrow} & \textbf{Timeouts \textdownarrow} \\
\hline
4 & 90.3\% & 9.7\% \\
\hline
5 & 86.7\% & 13.3\% \\
\hline
6 & 83.8\% & 16.2\% \\
\hline
7 & 82.3\% & 17.7\% \\
\hline
8 & 71.4\% & 28.6\% \\
\hline
3 Dynamic Obstacles & 81.4\% & 18.6\% \\
\hline
4 Dynamic Obstacles & 71.3\% & 28.7\% \\
\hline
\end{tabular} }
\caption{\textcolor{black}{Results for obstacle avoidance tests with varying numbers of obstacles of different shapes, including dynamic ones. While the robots successfully avoid collisions, the number of timeouts increases with obstacle density, reflecting the system’s conservative behavior and the challenges of navigating in more constrained environments.}}
\label{tab:more_obst}
\vspace{-20pt}
\end{table}


To evaluate the MPC's impact on agent training, we simulated two environments commonly used in the safe MARL literature \cite{zhang2022spatial,sheebaelhamd2021safe,cai2021safe,zhang2019mamps,zhang2022barrier}. The first environment is an empty walled setup to teach the robots to move the centroid to the goal. The second environment includes randomly placed cylindrical obstacles for collision avoidance learning. We train the policies in both environments with and without the MPC. We utilized the same network parameters to isolate the MPC filter's effects. Additionally, we incorporated a penalty in the RL policy for deviations from MPC-safe actions. \textcolor{black}{To maintain stable training despite the MPC's modified actions, we store the RL action in the replay buffer and include the MPC action as the previous action in the observation space.} Comparative metrics include average formation error, number of goals reached per episode, and number of collisions. Each episode terminates once the maximum number of steps is reached, or one of the robots collides or is stuck.

The average formation error is defined as follows: 
\vspace{-2pt}
\[
\begin{aligned}
\small
Formation\,Error = \frac{1}{N}\sum_{i=1}^N [dist_{i,j} - dist_{i,j}^{\mathit{ref}}]
\end{aligned}
\]
\vspace{-2pt}
where $dist_{i,j}^{\mathit{ref}}$ is the reference distance between robots $i$ and $j$, while $dist_{i,j}$ is the actual distance.

We denote the agents without the MPC filter as \textbf{ATT}, and those with the MPC as \textbf{ATT\_{MPC}}. The results, displayed in Fig. \ref{fig:filter}, highlight the efficacy of the MPC filter in training enhancement. Notably, \textbf{ATT\_{MPC}} reduces the average formation error to less than 0.5 meters within 1,000 episodes, a significant improvement over \textbf{ATT}, which requires about 4,000 episodes to achieve similar performance (Fig. \ref{fig:form_mpc}, \ref{fig:form2}). This enhancement is largely due to the MPC filter's ability to eliminate collisions, facilitating more effective exploration by the robots.

Additionally, Fig. \ref{fig:goals}, \ref{fig:goals2} illustrate that including the MPC-filter leads to reaching more goals per episode compared to the pure learning agents. These experiments show that we are able to decrease the number of episodes required to achieve the desired behavior, which is a crucial aspect for safe reinforcement learning, since fewer episodes means fewer resets of the environment in the real-world.
\vspace{-10pt}
\subsection{Testing Against Baselines in Simulation}
\label{sec:sims}

We evaluated our approach against three state-of-the-art baselines to validate its effectiveness in motion planning tasks. The baselines include:
\begin{itemize}
    \item \textbf{MASAC:} A multi-agent variant of the Soft Actor-Critic (SAC) \cite{he2022multiagent}, as it outperformed several MARL baselines in the motion planning tasks.
    \item \textbf{MADDPG:} The Multi-Agent Deep Deterministic Policy Gradient \cite{lowe2017multi}, a standard benchmark in MARL studies.
    \item \textbf{DMPC:} A decentralized model predictive control approach \cite{dawood2023handling}, using the centroid's reference as the goal for each robot, while maintaining inter-robot distances.
\end{itemize}

The RL agents were trained over two stages, each consisting of 4,000 episodes, with pre-trained agents loaded for further training in the second stage to adapt to obstacle avoiding scenarios. We compare between the agents in different scenarios.

\textbf{Reaching a Single Goal:} In the first scenario, we tested the ability of the robots to reach the desired goals starting from different configurations with and without obstacles in the environment. At the start of each episode, the locations of the robots, obstacles, and goal location are randomized. The results are summarized in Table \ref{tab:sims}. Our approach outperforms the baselines in terms of all three metrics. \textcolor{black}{Additionally, we evaluated our approach in more complex environments by introducing up to eight static obstacles, including cubes and wall sections, as well as dynamic obstacles, while training was limited to cylindrical obstacles. The robots avoided collisions in all cases, but timeouts increased with obstacle density, reflecting the system’s conservative behavior and the challenge of navigating constrained spaces. Results are in Table \ref{tab:more_obst}.}

\textbf{S-Shaped Path Following:} In this scenario, we simulated S-shaped paths for the centroid of the robot formation to replicate a real-world application where robots must navigate cooperatively following a path, e.g. conducting surveillance across multiple locations. We defined eight distinct S-shaped paths and initiated the robots from 15 varied starting configurations for each path.

\begin{figure*}[!t] \centering 
\rotatebox[origin=c]{90}{\bfseries Starting\strut}
\begin{subfigure} {0.185\linewidth} 	\centering 	\includegraphics[width=\linewidth]{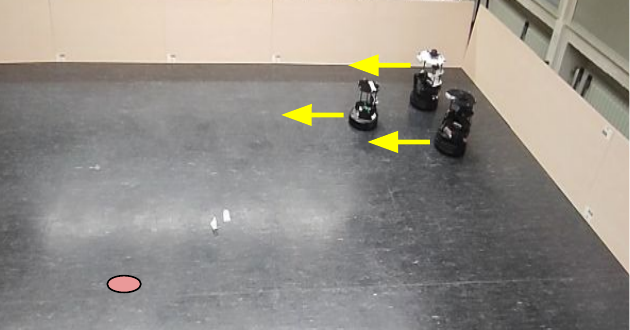} \subcaption{In-formation start} \label{fig:conf1a} \end{subfigure} 
\begin{subfigure} {0.185\linewidth} 	\centering 	\includegraphics[width=\linewidth]{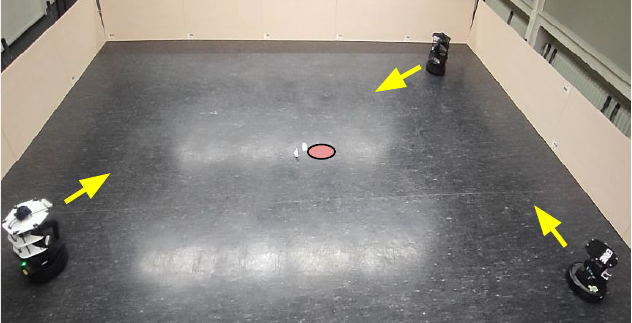} \subcaption{3-corners start} \label{fig:conf2a} \end{subfigure} 
\begin{subfigure} {0.185\linewidth} 	\centering 	\includegraphics[width=\linewidth]{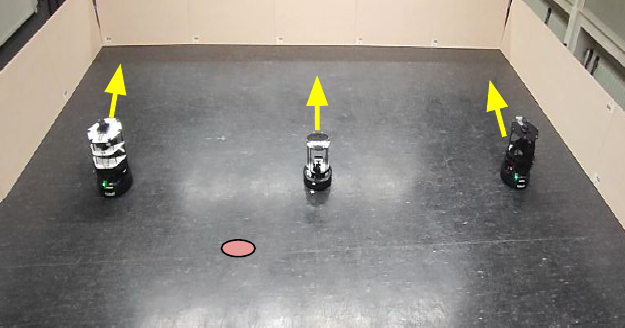} \subcaption{Centerline start} \label{fig:conf4a} \end{subfigure} 
\begin{subfigure} {0.185\linewidth} 	\includegraphics[width=\linewidth]{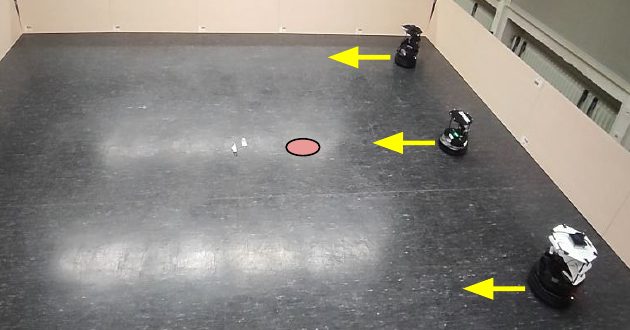} \subcaption{Collinear start} \label{fig:conf3a} \end{subfigure} 
\begin{subfigure} {0.185\linewidth} 	\centering 	\includegraphics[width=\linewidth]{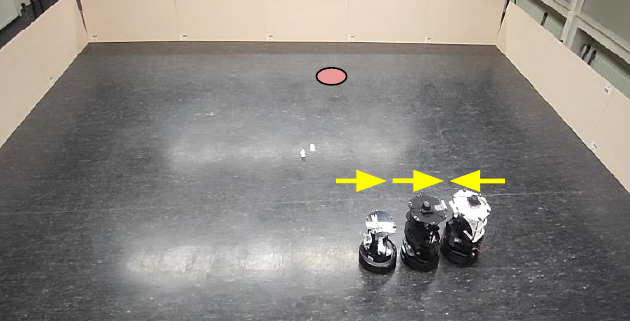} \subcaption{Face-to-face start} \label{fig:conf5a} \end{subfigure} \\[1.5ex] 
\rotatebox[origin=c]{90}{\bfseries Final\strut}
\begin{subfigure} {0.185\linewidth} 	\includegraphics[width=\linewidth]{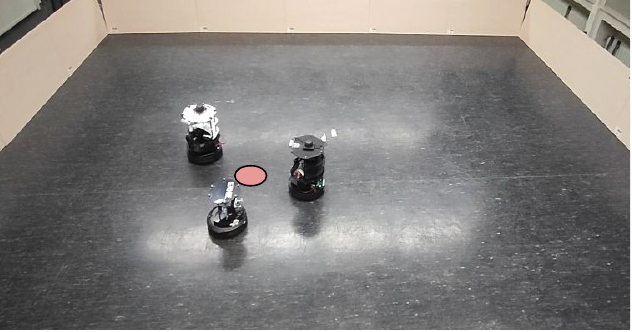} \subcaption{In-formation end} \label{fig:conf1b} \end{subfigure} 
\begin{subfigure} {0.185\linewidth} 	\centering 	\includegraphics[width=\linewidth]{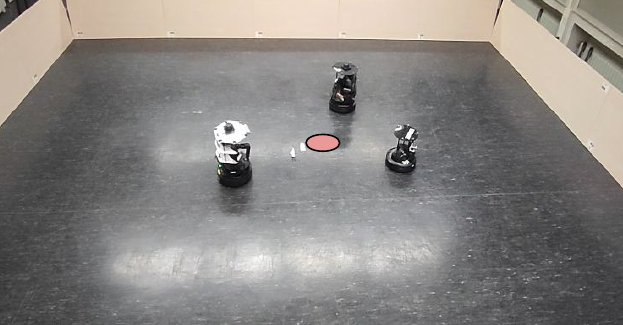} \subcaption{3-corners end} \label{fig:conf2b} \end{subfigure} 
\begin{subfigure} {0.185\linewidth} 	\centering 	\includegraphics[width=\linewidth]{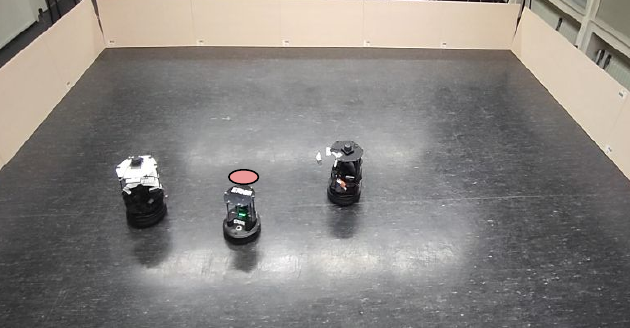} \subcaption{Centerline end} \label{fig:conf4b} \end{subfigure}
 \begin{subfigure} {0.185\linewidth} 	\centering 	\includegraphics[width=\linewidth]{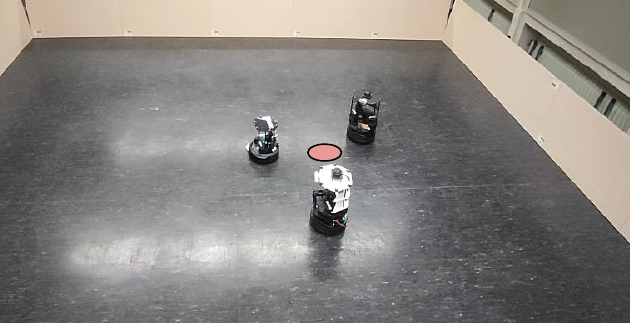} \subcaption{Collinear end} \label{fig:conf3b} \end{subfigure}
  \begin{subfigure} {0.185\linewidth} 	\centering 	\includegraphics[width=\linewidth]{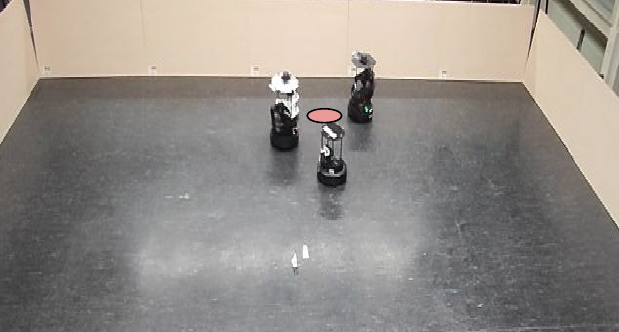} \subcaption{Face-to-face end} \label{fig:conf5b} \end{subfigure} \caption{Target Reaching Configurations test.
The top row indicates the starting configurations, while the bottom row shows the final reached configurations by our approach.
The yellow arrows indicate the starting orientation of the robots, while the red circles show the target location for the centroid of the formation.
The configurations are arranged from left to right in terms of difficulty.
The \textbf{Collinear} and \textbf{Facing each other} configurations were not experienced during the training, making them more difficult compared to the other configurations.
However, our approach was able to successfully complete the tasks with zero collisions.} 
 \label{fig:configs}
 \vspace{-11pt}
\end{figure*} 

The performance was evaluated based on completion time, collisions, and formation error, with results summarized in Table \ref{tab:sims}. Our approach demonstrated superior performance in completion time and collision avoidance, achieving the least time and zero collisions. While the DMPC showed the smallest formation error, it struggled with efficient positioning around the centroid, resulting in prolonged completion times. Among the learning-based methods, our approach exhibited the lowest formation errors, underscoring its effectiveness in maintaining formation while navigating.

\vspace{-8pt}	
\subsection{ Can we execute our method without the MPC? }

	\begin{figure}[!t]
	  \centering
	 \includegraphics[width=0.6\linewidth]{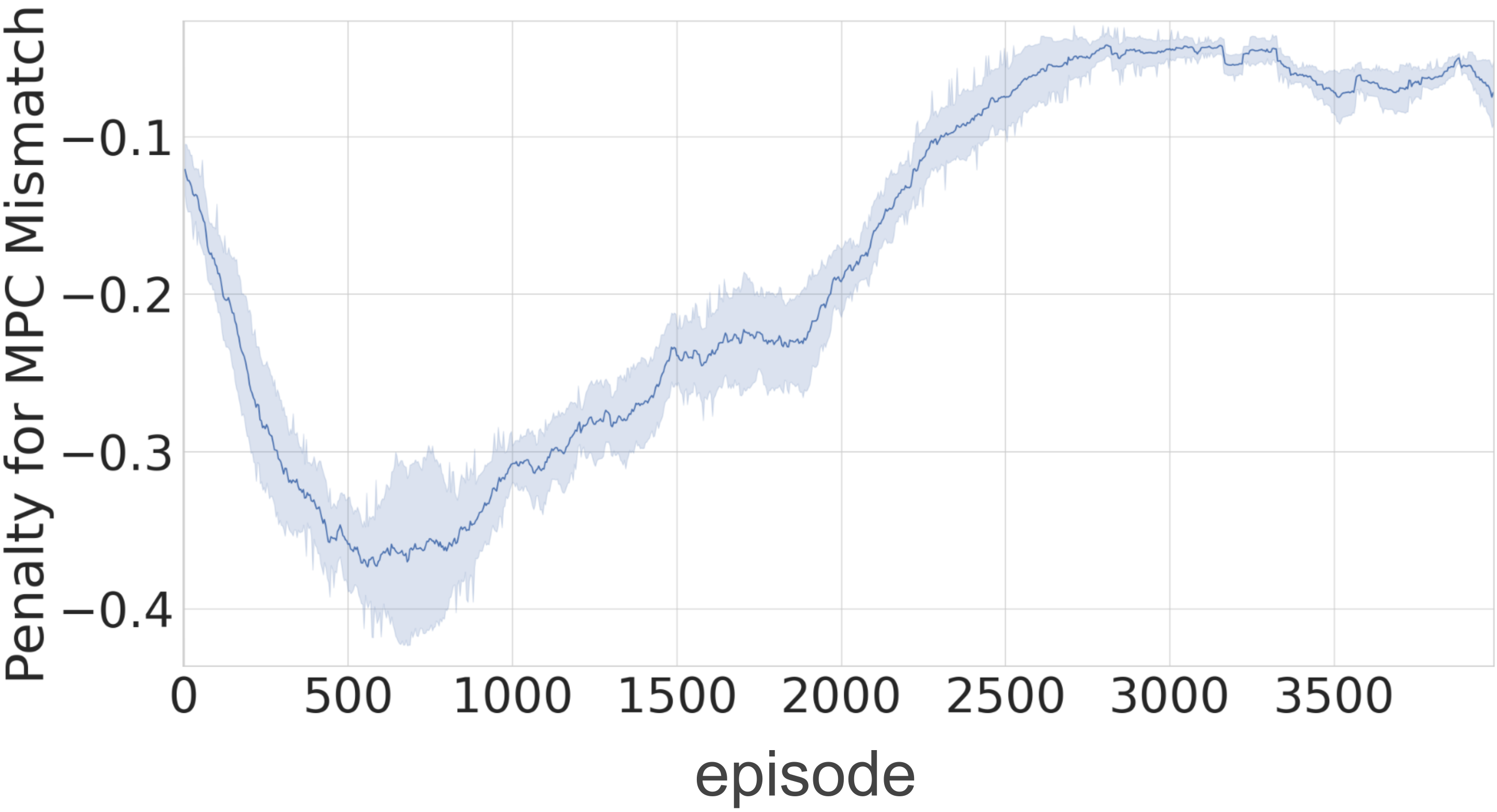}
	 \caption{The figure shows the penalty given to the agent for deviating from the MPC safe actions.The bold lines show the average while the shaded area show the standard deviation over three different random seeds. The penalty decreases through the training indicating that the agent learns to match the MPC safe actions.}  \label{fig:mpc_rews}
	  \vspace{-18pt}

	\end{figure}
\label{sec:ours_no_mpc}
In this section, we investigated if the MPC-filter can be disabled after training. We mentioned in Sec.\ref{sec:mpc_filter} that we added a penalty for deviating from the MPC-safe actions. In Fig. \ref{fig:mpc_rews}, we show that this penalty decreases over time, indicating the agents learn to adhere to the safe actions. To further test the agent without the MPC-safety filter, we used an agent previously trained with the MPC-filter, deactivated the MPC-safety filter and tested the agent in the same scenarios. Performance outcomes are detailed in \mbox{Table \ref{tab:sims}}. While the agent without the MPC-filter achieved over 90$\%$ in all tests showing that it can be used without the MPC, it failed to avoid collisions as effectively as it did with the MPC enabled, highlighting the MPC's critical role in ensuring safety.

\vspace{-8pt}
\subsection{Integrating the MPC Safety Layer with the Baselines: }
\label{sec:baselines_mpc}

We evaluated the baseline models after being retrained with the MPC safety layer and evaluated their performance using the established tests. The outcomes are detailed in \mbox{Table \ref{tab:sims}}. Incorporating the MPC with MASAC eliminated collisions, which increased the number of goals achieved but also led to more timeouts, indicating the MPC’s intervention to prevent collisions. Adding the MPC to MADDPG did not yield improvements, potentially due to the inherent stability issues of DDPG \cite{haarnoja2018soft}. The frequent timeouts suggest that the MPC often stops the robots to avoid collisions, resulting in timeouts due to the robots being stuck.
\begin{table}[!t]
\centering
\resizebox{8.2cm}{!}
{
\begin{tabular}{|c|c|c|c|c|}
\hline
\multicolumn{5}{|c|}{\textbf{Reaching a Single Goal}} \\
\hline
\textbf{Agent} & \multicolumn{2}{c|}{\textbf{Ours}} & \multicolumn{2}{c|}{\textbf{MASAC \cite{he2022multiagent}}} \\
\hline 
 {\textbf{Configuration}} & \textbf{Goals reached} \textbf{\textuparrow}  & \textbf{Colls} \textbf{\textdownarrow} & \textbf{Goals reached} \textbf{\textuparrow}  & \textbf{Colls} \textbf{\textdownarrow}\\ 
\hline
In-formation & 100\% &  \textbf{0} & 100\%  & 1.0 $\pm$ 0\\
\hline
3-corners & 100\%  &  \textbf{0}  &  100\%  & 1.0 $\pm$ 0 \\
\hline
Centerline & 100\%  &  \textbf{0} &  100\%  & 1.5 $\pm$ 0.7\\
\hline
Collinear & 100\%  & \textbf{0} &  100\%   & 0.5 $\pm$ 0.7\\
\hline
Face-to-face  &  100\%   & \textbf{0}  &  100\%  & 2.5 $\pm$ 0.7  \\
\hline
\end{tabular} }
\caption{\textcolor{black}{The table shows the performance of our approach against multi-agent SAC~(MASAC) in the real-world. Each scenario is tested two times using each approach. Our approach makes zero collisions in all configurations. The last two configurations were previously unseen during the training, nevertheless our approach was able to reach the targets with zero collisions unlike the MASAC.}}
\label{tab:configs}
\vspace{-21pt}
\end{table}

 \begin{figure*}[!t]
\centering
\begin{subfigure} {0.22\linewidth} 
  	\includegraphics[width=\linewidth]{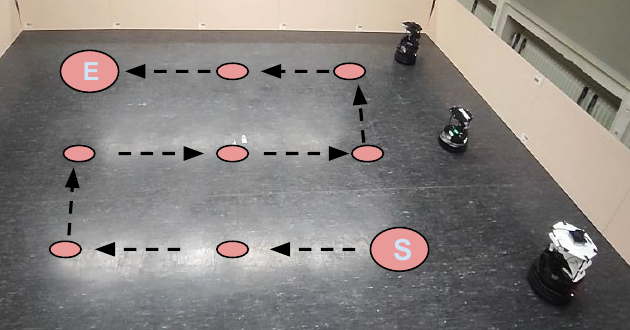}
  \subcaption{S-Shaped path}   \label{fig:a_spath}
    \end{subfigure}
  \begin{subfigure} {0.25\linewidth} 
  	\includegraphics[width=\linewidth]{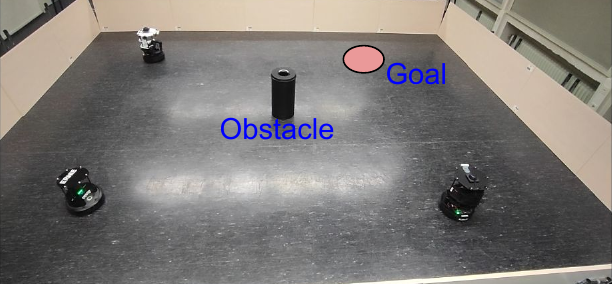}
  \subcaption{Scenario A}   \label{fig:b_obs}
    \end{subfigure}
  \begin{subfigure} {0.25\linewidth} 
  	\centering
  	\includegraphics[width=\linewidth]{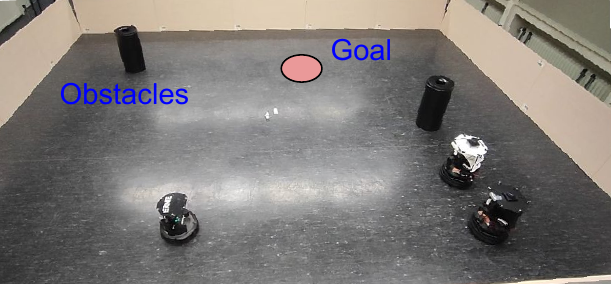}
  \subcaption{Scenario B}   \label{fig:c_obs}
    \end{subfigure}
  \caption{The figures show the S-Shaped path tests and two scenarios of the obstacle avoidance tests. \textbf{(a)} shows the S-shaped path. The red circles indicate the target locations for the centroid of the formation, where the first and final targets are indicated by the letters \textbf{S}, \textbf{E}, respectively. \textbf{(b,c)} show two different scenarios for obstacle avoidance. The robots start from the shown configurations and navigate towards the goal while avoiding collisions with the obstacles shown in the figures.}
 \label{fig:tests_other}
  \vspace{-8pt}

\end{figure*}

 \begin{table*}[!t]
\centering
\resizebox{13cm}{!}{
\begin{tabular}{|c|c|c|c|c|c|c|c|c|}
\hline
&\multicolumn{4}{c|}{\textbf{Goals Reaching With Obstacles}} &\multicolumn{4}{c|}{\textbf{S-Shaped Path}} \\
\hline
\makecell{\textbf{Agent}} & \makecell{Success}\textbf{ \textuparrow} & \makecell{Tout \textbf{\textdownarrow}} & \makecell{Colls \textbf{\textdownarrow}} & \makecell{Touts\textbf{\textdownarrow}}&  \makecell{Goals reached} \textbf{\textuparrow} &  \makecell{Completion time (sec.)}\textbf{\textdownarrow} & Collisions \textbf{\textdownarrow }& \makecell{Avg form err (m)}\textbf{\textdownarrow} \\
\hline 
Ours & \textbf{100\%} & 0 & \textbf{0} & \textbf{0} & \textbf{100\%} &  \textbf{143.2 $\pm$ 5.7} &  \textbf{0} & \textbf{0.34 $\pm$ 0.016} \\
\hline
MASAC \cite{he2022multiagent}& 0\% & 0 & 100\% & 0 &  96.2\% &  212.2 $\pm$ 53.8 &  7.3 $\pm$ 1.52 &  0.58 $\pm$ 0.35 \\
 \hline 
\end{tabular} }
\caption{The aggregated results for the obstacle avoidance and the S-Shaped path tests. Our approach completes all four tests with zero collisions. The MASAC was not successful in reaching the goals in case of obstacle avoidance tests due to collisions with the obstacles in all tests. \textbf{The S-Shaped path test} was carried out three times for each approach. The MASAC reached 96.2\% of the targets, since the robots failed to reach some targets within the allowed number of steps. During the tests, if the robots take more than 200 steps to reach a target, the next target in the path is given instead, and the previous target is counted as a timeout.}
\label{tab:config_obst}
\vspace{-14pt}
\end{table*}
\vspace{-10pt}
\subsection{Real-World Experiments}
\label{sec:real}
In the accompanying video, we demonstrated the behavior of initial (random) policies with and without the MPC filter on real robots, showcasing that our approach enables safe training on physical robots. However, training multiple robots in the real world is currently impractical due to time requirements. Our framework, although collision-free, requires 25 hours of simulation time, which translates to about 100 hours of real-world training.

Previous studies on safe MARL \cite{cai2021safe,zhang2022spatial,sheebaelhamd2021safe,elsayed2021safe,zhang2019mamps} have only evaluated final policies in simulations. In contrast, we validate our trained policies on real robots. The aim of these experiments is to demonstrate the zero-shot transfer of behavior-based navigation and the safety of our approach.

For comparison, we selected MASAC as a baseline due to its superior simulation performance over MADDPG and DMPC (Sec. \ref{sec:sims}), and trained both for the same number of episodes. The policies were then tested in real-world scenarios identical to those used in simulations.

\textbf{Reaching a Single Goal: }We tested five different starting configurations with a target location for the formation centroid, as shown in Fig. \ref{fig:configs}. Each configuration was tested twice for each policy, with results summarized in Table \ref{tab:configs}. Our approach consistently achieved zero collisions across all configurations, including unseen scenarios, demonstrating superior generalization and safety. Additionally, our method reached the targets faster than MASAC, as it better coordinates robot movements.

\begin{figure}[!t] \centering \begin{subfigure} {0.36\linewidth} 	\includegraphics[width=\linewidth]{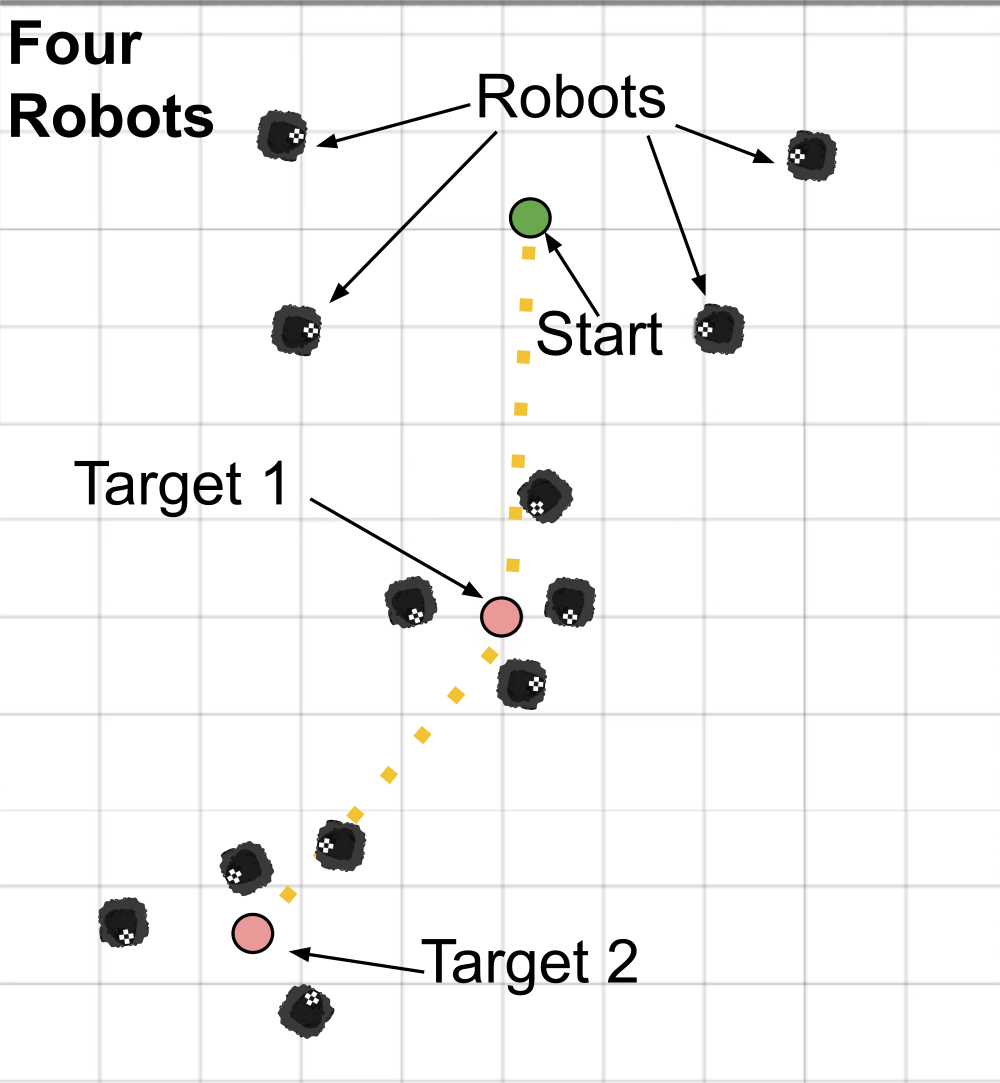} \subcaption{Testing with four agents} \label{fig:gener_4} \end{subfigure} \begin{subfigure} {0.36\linewidth} 	\centering 	\includegraphics[width=\linewidth]{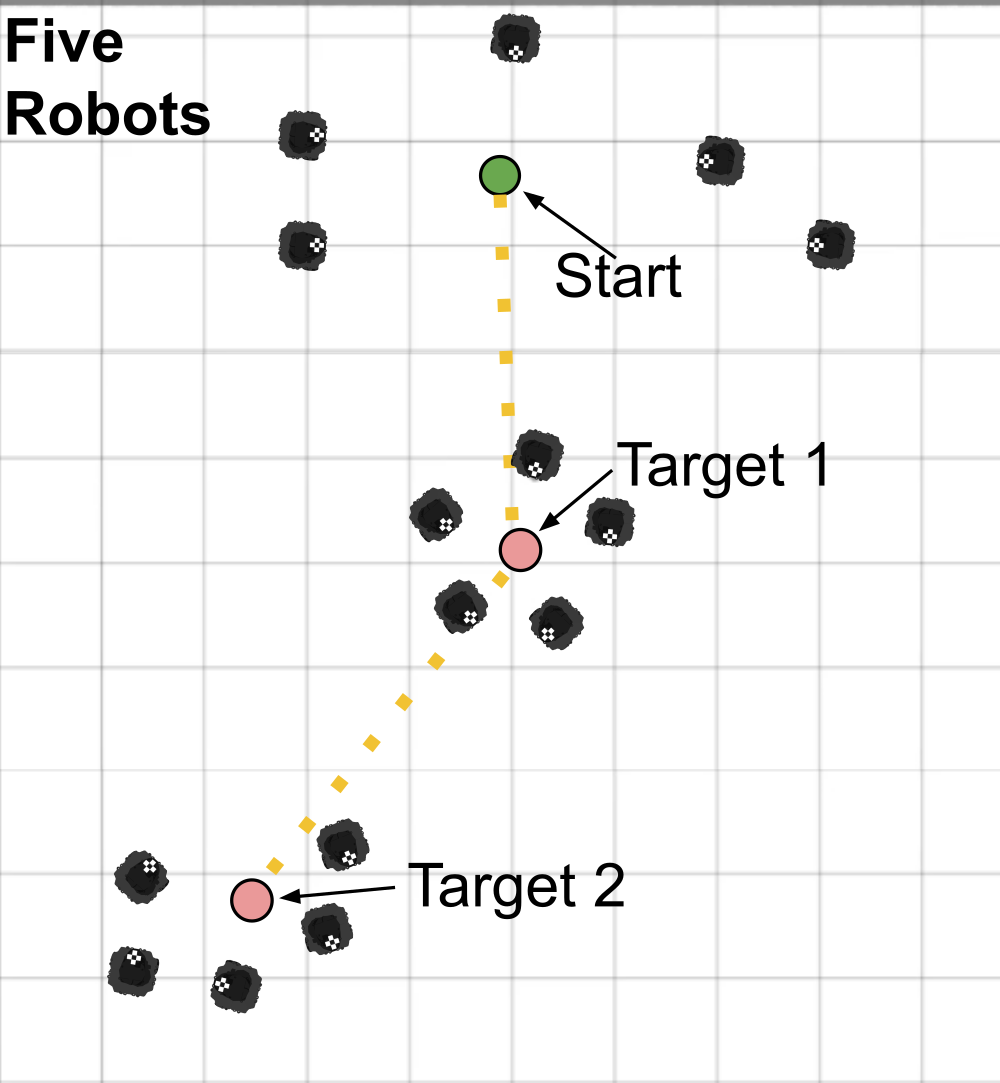} \subcaption{Testing with five agents} \label{fig:gener_5} \end{subfigure} 

\caption{Example scenarios \textbf{(a)} and \textbf{(b)} show the generalization of the trained policy to a larger team of robots. The robots successfully move their centroid from the starting location (green circle) to the target location (red circle).
The yellow lines indicate the movement of the centroid through the different locations.} 
 \label{fig:gener}
  \vspace{-9pt}

\end{figure} 
\begin{table}[!t]
\centering
\resizebox{7cm}{!}
{
\begin{tabular}{|c|c|c|c|c|c|}
\hline
\multicolumn{4}{|c|}{\textbf{Goals Reaching Test With More Robots}} \\
\hline
\multicolumn{1}{|c|}{\textbf{Number of Robots}} & \textbf{Goals reached }& \textbf{Collisions} & \textbf{Timeouts}\\ 
\hline
3 & 99\% &  0 & 1\%\\
\hline
4 & 97\% &  0 & 3\%\\
\hline
5 & 91\%  & 0 & 9\%\\
\hline
6 & 79\%  &  0 & 21\%\\
\hline
7 & 76\%  & 0 & 24\%\\
\hline
8 & 70\%  & 0 & 30\%\\
\hline

\end{tabular} }
\caption{The table shows the performance of our approach when generalizing to more robots, over 100 episodes. The percentage reflects the proportion of achieved goals, collisions, and timeouts in relation to the total number of trials. Our approach is able to generalize to more robots with zero collisions.
However, as the number of robots increased, we observed an increase in timeouts, attributed to the conservative nature of the MPC filter. 
}
\label{tab:gener}
 \vspace{-18pt}

\end{table}
In unseen scenarios, our approach required more maneuvers to avoid collisions, unlike MASAC, where robots reached the targets but failed to avoid collisions. For example, in the \textbf{face-to-face} configuration, our robots rotated and moved sequentially to avoid collisions, whereas MASAC robots collided and moved while touching, which explains the less time taken (these results are shown in the video). All tested configurations were reachable by both methods.
We further tested more configurations that were not reachable by the baseline in the video.
Both approaches were tested in obstacle avoidance scenarios. Fig. \ref{fig:tests_other} show two of these scenarios, and the results are shown in Table \ref{tab:config_obst}.

\textbf{S-Shaped Path Following: }In the S-shaped path scenario, Fig. \ref{fig:a_spath}, we run the test three times for each approach and recorded the number of collisions, the average formation error over the whole path, number of timeouts, as well as the time taken to complete the path.
The results are summarized in Table \ref{tab:config_obst}.
Our approach was able to successfully reach all targets without making collisions, and in less time compared to the MASAC. Additionally, our approach had less average formation error compared to the MASAC over all the runs.

 \vspace{-12pt}
\subsection{Generalizing to More Robots}
\label{sec:gener}

Finally, we show that by relying only on information from two neighbors, our method scales to more robots while maintaining a fixed observation size, regardless of the number of robots.

We evaluated the performance of the trained policy with up to eight robots, as shown in Fig. \ref{fig:gener}. For each robot, we defined two neighbors and tested the policy using the previously introduced goal-reaching scenarios. Metrics included the number of goals reached, collisions, and timeouts, with results summarized in Table \ref{tab:gener}. The robots successfully cooperated to move their centroid to the target locations, with zero collisions.
\textcolor{black}{However, as the number of robots increased, we observed an increase in timeouts, attributed to the conservative nature of the MPC filter and out-of-distribution scenarios introduced by additional robots. The policy was trained in an environment with only three robots and static obstacles, without exposure to dynamic obstacles. With more robots, they act as dynamic obstacles, leading to local congestion where robots block each other's paths, ultimately resulting in timeouts. These trade-offs are expected in safety-critical systems, where eliminating collisions is paramount. Addressing these limitations by incorporating dynamic obstacle scenarios into the training, along with integrating an oracle model to predict the future locations of nearby moving agents, represents a promising direction for future work to enhance the system's robustness and scalability.}

 \vspace{-10pt}
\section{Conclusions}
\label{sec:conclusion}

In this study, we presented a safe multi-agent reinforcement learning (MARL) approach to achieve behavior-based cooperative navigation without individual reference targets in real-world scenarios.
Our findings demonstrate that relying on the centroid of the formation is sufficient for robots to navigate cooperatively.
By integrating MARL, and model predictive control (MPC)-based safety filters, we ensured zero collisions during training and achieved faster convergence.
The inclusion of a safety layer not only eliminated collisions but also significantly improved training efficiency, leading to faster convergence of the MARL policies.
This has crucial implications for addressing the sim-to-real gap, as it highlights the benefits of incorporating safety layers into RL training.
Our trained policies, applied to real robots, outperformed baseline methods in various tests in terms of success rate, number of collisions and completion times, confirming the effectiveness of our approach.
Our experiments show that training on real robots is safe, eliminating collisions even during early exploration stages.

 \vspace{-10pt}
\bibliographystyle{IEEEtran}
\bibliography{bibliography}

\end{document}